\documentclass[runningheads]{llncs}
\usepackage{graphicx}
\usepackage{amsmath,amssymb} 
\usepackage{color}
\usepackage[width=122mm,left=12mm,paperwidth=146mm,height=193mm,top=12mm,paperheight=217mm]{geometry}
\usepackage{subcaption}
\usepackage[colorlinks=true]{hyperref}

\graphicspath{{images/}}
\newcommand{\etal}{\mbox{\emph{et al.\ }}}

\begin{document}
\pagestyle{headings}
\mainmatter

\title{Automatic Attribute Discovery with Neural Activations} 

\titlerunning{Automatic Attribute Discovery with Neural Activations}

\authorrunning{Sirion Vittayakorn \etal}

\author{Sirion Vittayakorn \inst{1}, Takayuki Umeda \inst{2}, Kazuhiko Murasaki \inst{2}, Kyoko Sudo \inst{2}, Takayuki Okatani \inst{3}, Kota Yamaguchi \inst{3}}
\institute{University of North Carolina at Chapel Hill, USA \and
NTT Media Intelligence Laboratories, Japan \and
Tohoku University, Japan
}

\maketitle

\begin{abstract}
How can a machine learn to recognize visual attributes emerging out of online community without a definitive supervised dataset? This paper proposes an automatic approach to discover and analyze visual attributes from a noisy collection of image-text data on the Web. Our approach is based on the relationship between attributes and neural activations in the deep network. We characterize the visual property of the attribute word as a divergence within weakly-annotated set of images. We show that the neural activations are useful for discovering and learning a classifier that well agrees with human perception from the noisy real-world Web data. The empirical study suggests the layered structure of the deep neural networks also gives us insights into the perceptual depth of the given word. Finally, we demonstrate that we can utilize highly-activating neurons for finding semantically relevant regions.

\end{abstract}

\section{Introduction}
In a social photo sharing service such as Flickr, Pinterest or Instagram, a new word can emerge at any moment, and even the same word can change its semantics and transforms our vocabulary set at any time. For instance, the word \textit{wicked} (literally means evil or morally wrong) is often used as a synonym of \textit{really} among teenagers in these recent years - \textit{``Wow, that game is wicked awesome!"}. In such a dynamic environment, how can we discover emerging visual concepts and build a visual classifier for each concept without a concrete dataset? It is unrealistic to manually build a high-quality dataset for learning every visual concepts for every application domains, even if some of the difficulty can be mitigated by the human-in-the-loop approach~\cite{biswas2013simultaneous,branson2010visual}. All we have are the observations but not definitions, provided in the form of co-occurring words and images.

In this paper, we consider an automatic approach to learn visual attributes from the open-world vocabulary on the Web. There have been numerous attempts of learning novel concepts from the Web in the past~\cite{6751285,ferrari2007learning,levan2014,Zhou_2015_CVPR,berg2010automatic}. What distinguishes our work from the previous efforts is in that we try to understand potentially-attribute words in terms of perception inside deep neural networks. Deep networks have demonstrated outstanding performance in object recognition~\cite{Krizhevsky_imagenetclassification,Simonyan2015,He2015}, and successfully applied to a wide range of tasks including learning from noisy data~\cite{TongXiao2015,Vo2015} or sentiment analysis~\cite{You2015,Jou2015}. In this paper, we focus on the analysis of neural activations to identify the degree of being visually perceptible, namely \textit{visualness} of a given attribute, and take advantage of the layered structure of the deep model to determine the semantic depth of the attribute.

We collect two domain-specific datasets from online e-commerce and social networking websites. We study in domain-specific data rather than trying to learn general concept on the Web~\cite{6751285,levan2014} to isolate contextual dependency of attributes to object categories. For example, the term \textit{red eye} can refer to an overnight airline flight or an eye that appears red due to illness or injury.
This contextual dependency can cause an ambiguity for the visual classifier (\textit{red} classifier); i.e., sense disambiguation. In this paper, we use a single-domain dataset to reduce such a semantic shift to study consistent meaning of attributes under the fixed context~\cite{Lampert2009,berg2010automatic}.
\begin{figure}[t]
  \centering
  \includegraphics[width=\columnwidth]{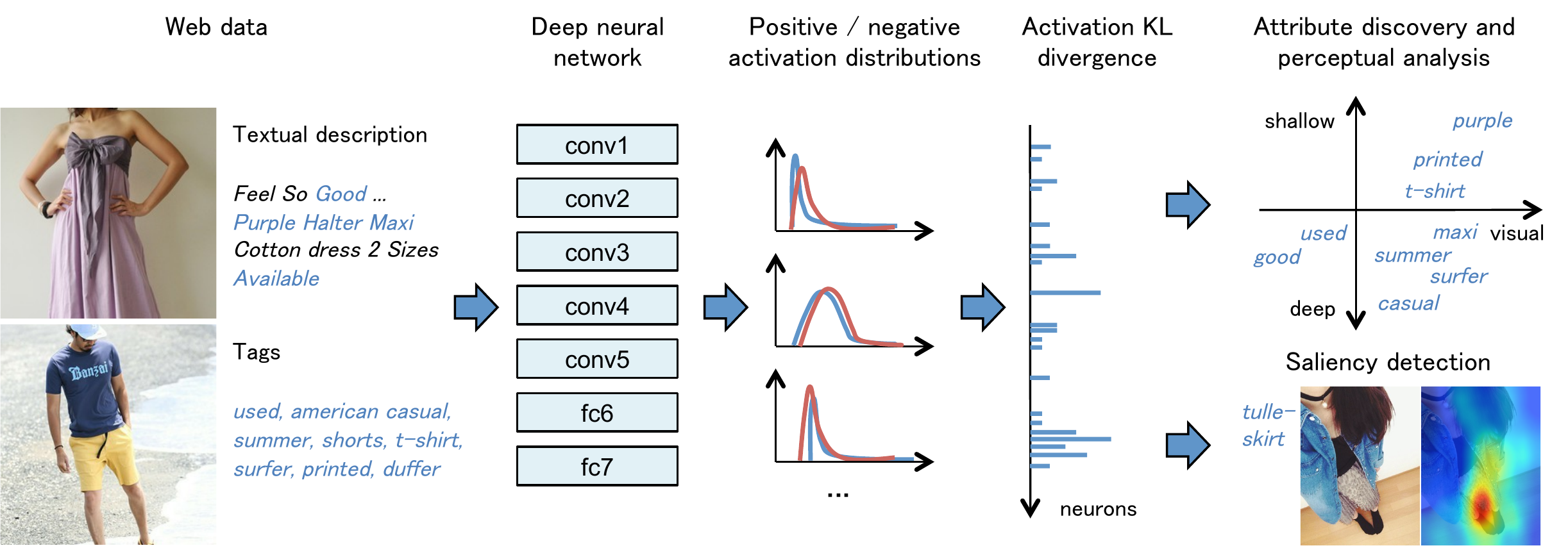}
  \caption{Our attribute discovery framework.}
  \label{fig:pipeline}
\end{figure}

We show that, using a trained neural network, we are able to characterize a visual attribute word by the divergence of neural activations in the weakly-annotated data. Figure \ref{fig:pipeline} illustrates our framework. Our approach starts by cleaning the noisy Web data to find potentially visual attributes in the dataset, then splits the data into positive and negative sets. Using a pre-trained neural network, we identify highly activating neurons by KL divergence of activations. We show that we can use the identified neurons (prime units) for 1) learning a novel attribute classifier that is close to human perception, 2) understanding perceptual depth of the attribute, and 3) identifying attribute-specific saliency in the image. We summarize our contributions in the following.
\begin{enumerate}
  \setlength\itemsep{0em}
  \item We propose to utilize the divergence of neural activation as a descriptor to characterize visual concept in the noisy weakly annotated dataset. The neurons identified by the divergence can help learn a visual attribute classifier that has a close proximity to human perception.
  \item We empirically study the relationship between human perception and the depth of activations to understand the visual semantics of attribute words.
  \item We show that the highly activating neurons according to the divergence are also useful for detecting attribute-specific saliency in the given image.
  \item We collect two noisy datasets from the Web to evaluate our framework. The empirical study shows we are able to learn a domain-specific visual attributes without manual annotation.
\end{enumerate}

\section{Related work}

\subsubsection{Attribute discovery}
Our work is related to the recent work on concept discovery from a collection of images from the Web~\cite{6751285,ferrari2007learning,levan2014,Zhou_2015_CVPR,sun2015automatic}. Early work by Ferrari \etal~\cite{ferrari2007learning} learns visual models
of given attributes (e.g., red, spotted) from images collected from text search. NEIL~\cite{6751285} aims at discovering common sense knowledge from the Web starting from small exemplar images per concept. LEVAN~\cite{levan2014} starts from mining bi-gram concepts from a large text corpus, and automatically retrieves training images from the Internet and learn a full-fledged detector for each concept. ConceptLearner~\cite{Zhou_2015_CVPR} uses weakly labeled image collections from Flickr to train visual concept detectors. Shankar \etal~\cite{shankar2015deep} study the attribute discovery in weakly-supervised scenario, where the goal is to identify co-occuring but missing attributes in the dataset while learning a deep network. Recent work by Sun \etal~\cite{sun2015automatic} takes advantage of natural language to discover concepts for retrieval scenario. The automatic attribute discovery by Berg \etal~\cite{berg2010automatic} is close to our work in that the work tries to evaluate visualness of the discovered synsets of attributes in the e-commerce scenario. The major difference of our approach from the previous works is that we aim at discovering attribute words and also characterizing the attribute perception using neural activations.

\subsubsection{Neural representation} Thanks to the outstanding performance of deep neural networks in various tasks such as object recognition~\cite{Krizhevsky_imagenetclassification,43022,Simonyan2015,He2015} or domain adaptation for visual recognition task~\cite{Oquab2014}, the deep analysis of the intermediate representation of the neural networks has been getting more attention~\cite{yosinski2015understandingCNN,Zeiler2014,Zhou2014}. Escorcia \etal~\cite{Escorcia2015} and Ozeki \etal~\cite{ozeki2014understanding} study the relationship between neural representation and attributes. In this paper, we aim at utilizing the intermediate representation for visually characterizing unknown words in the noisy dataset, and study how the representation relates to human perception of attributes.

\subsubsection{Class-specific saliency detection}
Detecting class or attribute-specific saliency has been studied in the past in various forms, for example, as co-segmentation~\cite{Chai2012}, parts~\cite{Oquab2014,Simon2015} or latent parts discovery~\cite{Kiapour2014}, and weakly-supervised ~\cite{Guillaumin2014,zhou2015cnnlocalization} or fully-supervised labeling~\cite{pan16fastsaliency,panWJ15}. While Simonyan \etal~\cite{Simonyan2014} uses gradient as a class-specific saliency, we demonstrate that the receptive field of neurons~\cite{Zhou2014} can effectively identify the attribute-specific regions with our activation divergence. Our neuron-level saliency detection performs comparable to gradient-based approach~\cite{Simonyan2014}, and can also reveal us insight into how learning changes the neurons' response to visual stimuli.

\section{Datasets and pre-processing}
\label{sec:datasets}

\subsection{Etsy dataset}\label{sec:etsy_dataset}
Etsy dataset is a collection of data from the online market of handcrafted products. Each product listing in Etsy contains an image, a title, a product description, and various metadata such as tags, category, or price. We initially crawl over 2.8 million product pages from \href{https://www.etsy.com}{etsy.com}. Considering the trade-off between dataset size and domain specificity, we select the product images under the clothing category, which include 247 subcategories such as {\tt clothing/women/dress}.


\subsubsection{Near-duplicate removal}
As common in any Web data, the raw data from Etsy contain a huge amount of near-duplicates. The major characteristics of Etsy data are the following: There are many shops, but the number of sold items per shop exhibits a long-tail. The same shop tends to sell similar items, e.g., the same black hoodie in the same background with a different logo patch, and in an extreme case, just a few words (proper nouns) are different in the product description. Our near-duplicate removal is primarily designed to prevent such proper nouns from building up a category. We observe that without the removal, we severely suffer from overfitting and end up with meaningless results.

Based on the above observation, we apply the following procedure to remove near-duplicates in Etsy: 1) Group product listings by shop. 2) Compute a bag-of-words from title and description except English stop words for each item within the group. 3) Compute the cosine distance between all pairs of products. 4) Apply agglomerative clustering by thresholding the pairwise cosine distance. 5) Randomly pick one product from each cluster.
We apply the duplicate-removal for all shops in the dataset, and for each shop we merge any pairs of product having less than $0.1$ cosine distance into the same cluster. After the near-duplicate removal, we observed that roughly 40\% of the products in Etsy were considered near-duplicates. We obtained 173,175 clothing products for our experiment.

\subsubsection{Syntactic analysis}\label{sec:syntactic_analysis}
Given the title and description of the product in Etsy dataset, we apply syntactic analysis~\cite{Marneffe06generatingtyped} and extract part-of-speech (POS) tags for each word. In this paper, we consider 250 most frequent adjectives ({\tt JJ}, {\tt JJR}, and {\tt JJS} tags) as potential attribute words. Unless noted, we use the (50\%, 25\%, 25\%) splits for train, test, and validation in the following experiments.


\subsection{Wear dataset}
We crawled a large collection of images from the social fashion sharing website, \href{http://wear.jp}{wear.jp}. The website contains an image, associated shots from different views, list of items, blog text, tags, and other metadata. The images in Wear dataset are extremely noisy; Many users take a photo with a mobile device under uncontrolled lighting conditions and inconsistent photo composition, making it very challenging to apply any existing fashion recognition approach~\cite{yamaguchi2015retrieving}. From the crawled data, we use the random subset of 212,129 images for our experiments.

\subsubsection{Merging synonyms and translations}
From Wear dataset, we select user-annotated tags for candidate words. The majority of tags from Wear dataset are written in Japanese (some in English), but there are also multiple synonyms treated as different tags including typos. We observe such synonyms and translations creating many duplicates.
To mitigate this problem, we remove synonyms by translating all words to English, using Google Translate, and merge words that maps to the same English word. After translation, we pick up the most frequent 250 tags as a set of attribute candidates and use for our experiment. Note that machine-translation is not perfect and we manually fix translation errors in the selected tags. 

\section{Attribute discovery}
Our attribute discovery framework starts by first splitting the weakly-annotated dataset into positive and negative sets, then computes Kullback-Leibler divergence (KL) for each activation unit in the deep neural network. We use the KL divergence to determine the important neurons for the given attributes. With these selected neurons, we can estimate the degree of \textit{visualness} of attributes.

\subsection{Divergence of neural activations}
Although the image representation (neural activations) from the deep network captures numerous discriminative features in an image~\cite{Zeiler2014}, each neuron in the network only sparsely responds to visual stimuli. We attempt to find neurons that highly respond to the visual pattern associated with a given attribute word. We propose to use the KL divergence of activations to identify these highly responding neurons or \textit{prime units} for the given attribute.

Our framework starts by splitting the dataset $D$ into positive and negative sets according to the weak annotation (adjectives or tags in Sec \ref{sec:datasets}). Positive or negative sets $D^+_u$, $D^-_u$ are images with or without the candidate attribute-word $u$. Note the noisy annotation contains both false-positive and false-negative samples. 
Using a pre-trained neural network, we compute the empirical distribution of neural activations from all of the units in the network. Let us represent the empirical distribution of the positive / negative set by $P^+_i$ and $P^-_i$ for each neuron $i$. For convolutional layers, we max-pool the spatial dimension in all channels and compute histograms $P^+_i, P^-_i$, since the maximum response is sufficient to identify unique units regardless of the location. Finally, we compute the symmetric KL divergence $S_i$ for each activation unit $i$ of the network:
\begin{eqnarray}
S_i(u|D) & \equiv & D_{\text{KL}}(P^+_i||P^-_i) + D_{\text{KL}}(P^-_i||P^+_i) \nonumber \\
& = & \sum_{x}P^+_i(x)\log\frac{P^+_i(x)}{P^-_i(x)} + \sum_{x}P^-_i(x)\log\frac{P^-_i(x)}{P^+_i(x)},
\label{eq:kl_divergence}
\end{eqnarray}
where $x$ is the activation of the unit corresponding to histogram bins.
The resulting KL divergence $S_i(u|D)$ serves as an indicator to find prime units for the word $u$. The intuition is that if the word is associated to specific visual stimuli, the activation pattern of the positive set should be different from the negative set and that should result in a larger KL divergence for visual attributes (e.g., red, white, floral, stripped) than less visual attributes (e.g., expensive or handmade). In other word, we should be able to identify the visual pattern associated to the given word by finding neurons with higher KL divergence.

\subsection{Visualness}
We follow the previous work~\cite{berg2010automatic} and define the visualness in terms of the balanced classification accuracy given the positive and negative sets:
\begin{equation}
V(u|f) \equiv \text{accuracy} (f, D^+_u, D^-_u),
\label{eq:visualness}
\end{equation}
where $f$ is a binary classification function. To eliminate the bias influence in the accuracy, we randomly subsample the positive and negative sets $D^+_u, D^-_u$ to obtain balanced examples (50\%-50\%). We use neural activations as a feature representation to build a classifier, and use the KL divergence $S_i$ as resampling and feature-selection criteria to identify important features for a given word $u$.

\subsubsection{Selecting and resampling by activations}
\label{sec:naive_bayes_subsampling}

The noisy positive and negative sets $D^+, D^-$ bring undesirable influence when evaluating the classification accuracy of the word (eq \ref{eq:visualness}). Here, we propose to learn a visual classifier in two steps; we first learn an initial classifier based only on the activations from prime units, then we rank images based on the classification confidence. After that, we learn a stronger classifier from the confident samples using all of the activations in the network. More specifically, we first select 100 prime units according to the KL divergence (eq \ref{eq:kl_divergence}), and use the activations from these units as a feature (100 dimensions) to learn an initial classifier\footnote{Gaussian Naive Bayes also works in our setting, but a stronger classifier such as SVM with RBF kernel tends to overfit.} using logistic regression~\cite{REF08a} and identify the confident samples for the second classifier. 


\subsubsection{Learning attribute classifier}
\label{sec:attribute_classifier}

Once we learn the initial classifier, we rank images based on the confidence, resample the same number of images from both positive and negative sets according to the ranked order, and learn another attribute classifier using logistic regression from all of the activations (9,568 dimensions). Although more than 8,000 activations are from FC layers, the information gain is not necessarily proportional to the number of dimensions; FC layers tend to fire only a handful neurons, whereas convolutional layers after max-pooling give dense activations. 
Finally, the accuracy evaluation (eq. \ref{eq:visualness}) on the balanced test set gives the visualness of the given word.

\subsection{Human perception}
\label{sec:human_perception}

To evaluate our approach, we collect human judgment of visualness using crowdsourcing, and compute the correlation between our visualness and human perception. Following the observation in~\cite{parikh2011relative} that it is harder for humans to provide the absolute visualness score of attribute than the relative score. Thus, we design a task on Amazon Mechanical Turk as follows; given a word, we provide two images to the annotators where one is from the positive set and the other is from the negative set. We ask annotators to pick an image that is more relevant to the given attribute, or if there is none, answer none. We pick the 100 most frequent words in Etsy dataset for evaluation purpose. For each word, we randomly pick 50 pairs of positive and negative images, and asked 5 annotators to complete one task. We define the human visualness $H(u)$ of word $u$ as the ratio of positive annotator agreements:
\begin{equation}
H(u) \equiv \frac{1}{N} \sum_k^N \vec{1}\left[ h_k^+(u) > \theta \right],
\label{eq:human_agreement}
\end{equation}
where $\vec{1}$ is an indicator function, $h_k^+(u)$ is the number of positive votes for image pair $k$, $N$ is the number of annotated images, and $\theta$ is a threshold. We set $\theta = 3$ for 5 annotators in our experiment. Eq. \ref{eq:human_agreement} allows us to convert the relative comparison into agreement score, which is in absolute scale.

\subsection{Experimental results}
\label{sec:visualness_experiments}

We use the Etsy dataset to evaluate our visualness.\footnote{\scriptsize Due to the translation issues, we were not able to get reliable human judgments in Wear dataset.}
Since neurons activate differently depending on the training data, we compare the following models:
\begin{itemize}
\item {\bf Pre-trained:} Reference CaffeNet model~\cite{Krizhevsky_imagenetclassification} implemented in~\cite{jia2014caffe}, pre-trained on ImageNet 1000 categories.
\item {\bf Attribute-tuned:} A CNN fine-tuned to directly predict the weakly-annotated words in the dataset, ignoring the noise. We replaced the soft-max layer in CaffeNet with a sigmoid to predict 250 words (Sec \ref{sec:etsy_dataset}).
\item {\bf Category-tuned:} A CNN fine-tuned to predict the 247 sub-categories of clothing using metadata in Etsy dataset, such as t-shirt, dress, etc.
\end{itemize}

We choose the basic AlexNet to evaluate how fine-tuning affects in our attribute discovery task, but we can also apply a different CNN architecture such as VGG~\cite{Simonyan2015} to do the same. The category-tuned model is to see the effect of domain transfer without overfitting to the target labels. We compare the following different visualness definitions against human.
\begin{itemize}
\item{\bf CNN+random:} Randomly subsample the same number of positive and negative images, learn a logistic regression from all of the neural activations (9,568 dimensions) in CNN, and use the testing accuracy to define the visualness. This is similar to the visualness prediction in the previous work~\cite{berg2010automatic} except that we use neural activations as a feature.
\item{\bf CNN+initial}: Testing accuracy of the initial classifier trained on the most activating neurons or prime units.
\item{\bf CNN+resample:} Testing accuracy of the attribute classifier trained on the resampled images according to the confidence of the initial classifier and learned from all of the neural activations, as described in Sec \ref{sec:attribute_classifier}.
\item{\bf Attribute-tuned:} Average precision of the direct prediction of the Attribute-tuned CNN in the balanced test set. We choose average precision instead of accuracy due to severe overfitting to our noisy training data.
\item{\bf Language prior:} The n-gram frequency of adjective-noun modification for the given attribute-word from the Google Books N-grams~\cite{37388}. We show the language prior as a reference to understand the scenario when we do not have access to visual data at all. The assumption is that for each of the object category in Etsy, visual modifier should co-occur more than non-visual words. We compute the prior using the sum of n-gram probability on attribute-category modification to 20 nouns in the Etsy clothing categories.
\end{itemize}

\subsubsection{Quantitative evaluation}
\begin{table}[t]
  \centering
  \caption{Visualness correlation to human perception.}
  \label{tab:human_correlation}
  \scriptsize
  \begin{tabular}{lccc}
  \hline
  Method & Feature dim. & Pearson & Spearman \\
  \hline
  Pre-trained+random (baseline) & 9,568 & 0.737 & 0.637 \\
  Pre-trained+initial & 100 & 0.760 & 0.663 \\
  Pre-trained+resample & 9,568 & \bf 0.799 & 0.717 \\
  \hline
  Attribute-tuned & 4,096 & 0.662 & 0.549 \\
  Attribute-tuned+random & 9,568 & 0.716 & 0.565 \\
  Attribute-tuned+initial & 100 & 0.716 & 0.603 \\
  Attribute-tuned+resample & 9,568 & 0.782 & \bf 0.721 \\  
  \hline
  Category-tuned+random & 9,568 & 0.760 & 0.684\\
  Category-tuned+initial & 100 & 0.663 & 0.480 \\
  Category-tuned+resample & 9,568 & 0.783 & 0.704 \\
  \hline
  Language prior & - & 0.139 &0.032 \\
  \hline
  \end{tabular}
\end{table}

Table \ref{tab:human_correlation} summarizes the Pearson and Spearman correlation coefficients to human perception using different definitions of visualness together with the feature dimension. \emph{Note that achieving the highest accuracy in classification does not mean the best proximity to human perception in the noisy dataset}. The results show that even though the initial classifiers learn from only 100 dimensional feature from prime units, they achieve the higher Spearman correlation to human perception than the random baselines with much larger feature. Moreover, resampling images by the initial classifier confidence improves the correlation to human perception over the random baseline in all models. These results confirm that feature-selection and resampling using the high-KL neurons help discovering visual attributes in the noisy dataset.


The result also suggests directly fine-tuning against the noisy annotation can harm the representational ability of neurons. We suspect that fine-tuning to a domain-specific data with possibly non-visual word leads to overfitting and suppresses neurons' activity even if they are important in recognition. 
The pre-trained network gives the slightly higher Pearson correlation perhaps because the neurons are trained on wider range of visual stimuli in the ImageNet than in a domain-specific data like Etsy, and that somehow helps reproducing human perception. The low correlation from language prior indicates the difficulty of detecting visual attributes only from textual knowledge.

\subsubsection{Qualitative evaluation}
Table \ref{tab:visual_words} lists the most and least visual attributes for selected methods. Note that the error in syntactic analysis incorrectly marked some nouns as adjective, such as {\it url} or {\it flip} (flip-flops) here. Generally, CNN-based methods result in a similar choice of the words. Language prior is picking very different vocabulary perhaps due to the lack of domain-specific knowledge in Google Books.

\begin{figure}[t]
  \centering
  \includegraphics[width=\columnwidth]{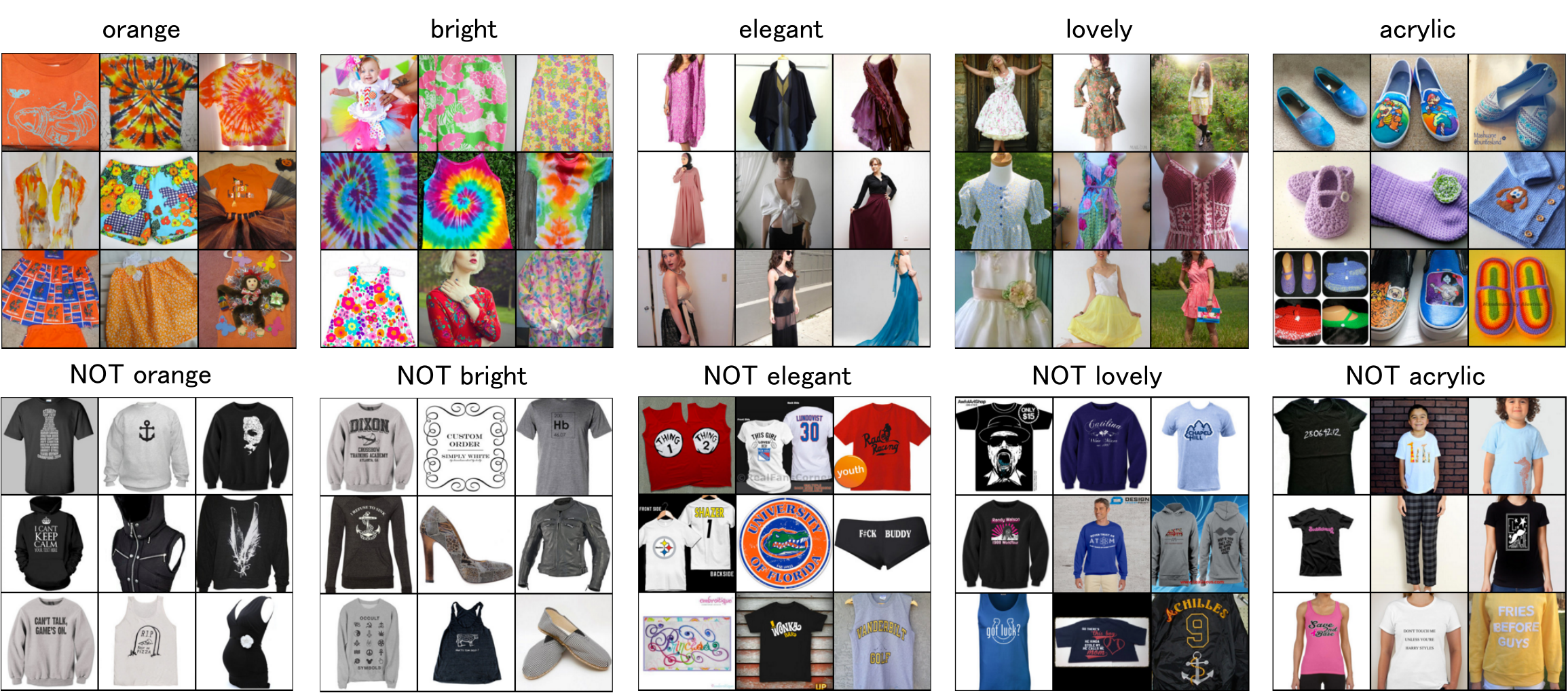}
  \caption{Examples of most and least predicted images for some of the attributes.}
  \label{fig:predicted_images}
\end{figure}

\begin{figure}[t]
  \centering
  \includegraphics[width=.6\columnwidth]{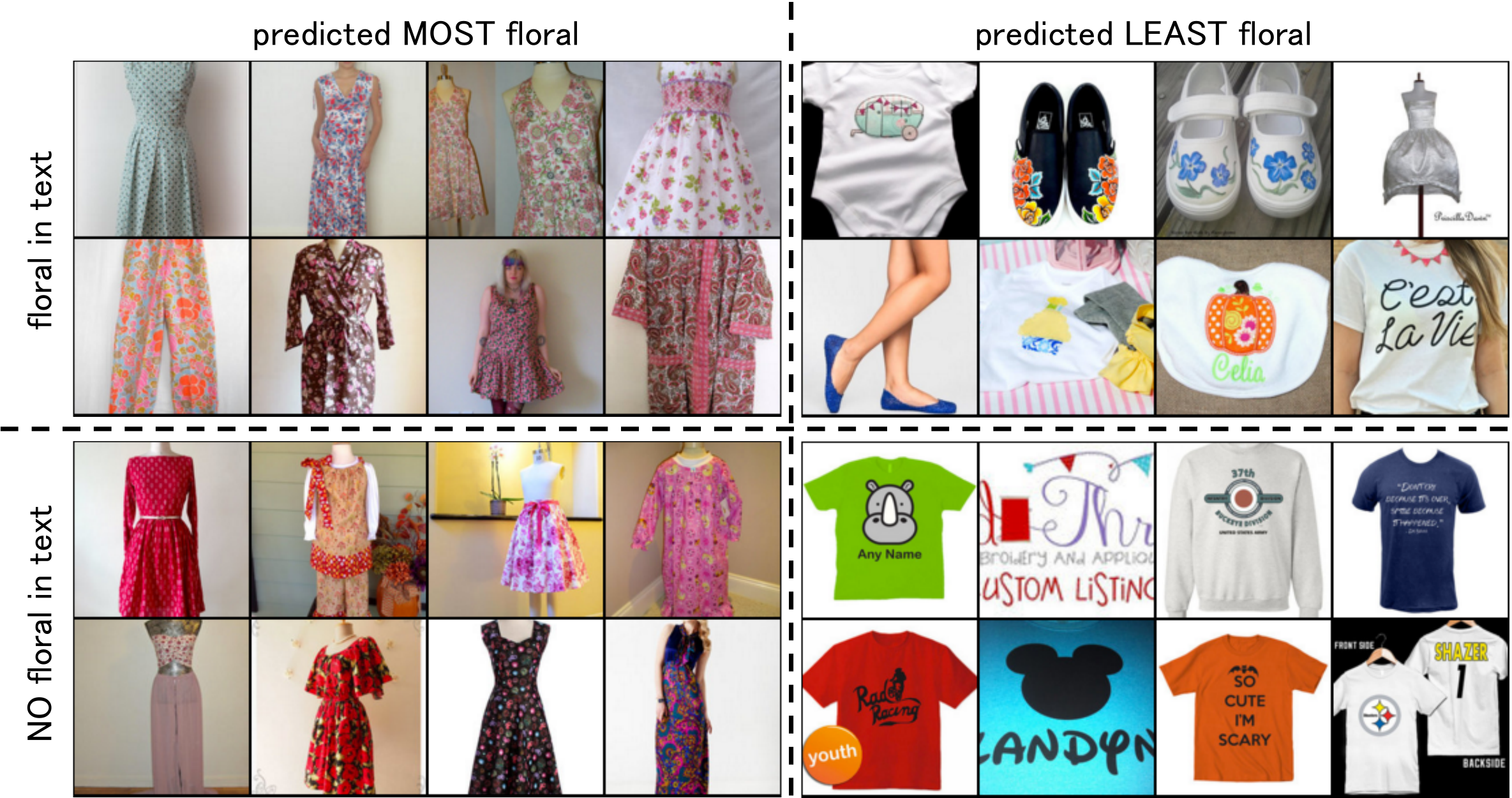}
  \caption{Most and least {\it floral} images. With our automatically learned classifier, we can discover false-negatives and false-positives in the dataset.}
  \label{fig:floral_prediction}
\end{figure}

Figure \ref{fig:predicted_images} shows examples of the most or least confident images according to the pre-trained+resample model. From concrete concepts like {\it orange} to more abstract concepts {\it elegant}, we confirm that our automatic approach can learn various attributes only from the noisy dataset. Figure \ref{fig:floral_prediction} shows examples of the most and least {\it floral} images from both positive and negative sets. The noise in the dataset introduces a lot of false-negatives (not mentioned but actually floral product) and false-positives (mentioned floral in text but not relevant to the product). Our automatically learned attribute classifiers can function as a dataset purifier in a noisy dataset.

\begin{table}[t]
  \centering
  \caption{Most and least visual attributes discovered in Etsy dataset.}
  \label{tab:visual_words}
  \scriptsize
  \begin{tabular}{lll}
  \hline
  Method & Most visual & Least visual\\
  \hline
  Human & flip pink red floral blue & url due last right additional \\
        & sleeve purple little black yellow & sure free old possible cold \\
  \hline
  Pre-trained+resample & flip pink red yellow green & big great due much own \\
                   & purple floral blue sexy elegant & favorite new free different good \\
  \hline
  Attribute-tuned & flip sexy green floral yellow & right same own light happy \\
                  & pink red purple lace loose & best small different favorite free \\
  \hline
  Language prior & top sleeve front matching waist & organic lightweight classic gentle adjustable \\
                 & bottom lace dry own right & floral adorable url elastic super \\
  \hline
  \end{tabular}
\end{table}


\section{Understanding perceptual depth}
\label{sec:perceptual_depth}
In this section, we explore how each layer in the neural networks relates to attributes.
It is well-known that neurons in a different layer activate to different types of visual pattern~\cite{Zeiler2014,Escorcia2015}. We further attempt to understand what type of semantic concepts directly relate to neurons using the KL divergence.

We consider the activation with respect to the layer depth. We compute the relative magnitude of max-pooled KL divergence for layer $l$:
\begin{eqnarray}
S_l(u|D) & \equiv & \frac{1}{Z} \max_{i \in l} S_i(u|D), ~ \text{where} ~ Z \equiv \sum_{l} \max_{i \in l} S_i(u|D).
\label{eq:layerwise_kl}
\end{eqnarray}
We are able to identify the most {\it salient words} by ranking attribute vocabulary based on $S_l(u|D)$. In the following experiments, we use 7 layers in CaffeNet.

We use both Etsy and Wear dataset for finding salient words at each layer. Table \ref{tab:layer-words} lists the most salient words for each layer of the pre-trained CNN in the two datasets. We can clearly see that more primitive visual concepts like color (e.g., {\it orange}, {\it green}) appear in the earlier stage of the CNN, and as we move down the network towards the output, we observe more complex visual concepts. We can observe the same trend from both Etsy and Wear datasets even though the two datasets are very different.
Note that there are non-visual words in a general sense due to the dataset bias. For example, {\it genuine} in Etsy tends to appear in the context of {\it genuine leather}, and {\it many} appear in the context of {\it many designs available} for sweatshirt products. Such dataset bias results in higher divergence of neurons' activity. One approach to deal with such context-dependency is probably to consider a phrase instead of a word.

\begin{table}[t]
  \centering
  \scriptsize
  \caption{Most salient words for each CNN layer.}
  
  \label{tab:layer-words}
  \begin{subtable}{\columnwidth}
    \centering
    \caption{Etsy dataset}
    \begin{tabular}{|p{.132\columnwidth}|p{.132\columnwidth}|p{.132\columnwidth}|p{.132\columnwidth}|p{.132\columnwidth}|p{.132\columnwidth}|p{.132\columnwidth}|}
    \hline
      \bf norm1      & \bf norm2      & \bf conv3      & \bf conv4      & \bf pool5      & \bf fc6        & \bf fc7        \\
    \hline
      orange        & green        & bright       & flattering   & lovely       & many         & sleeve         \\
      colorful      & red          & pink         & lovely       & elegant      & soft         & sole           \\
      vibrant       & yellow       & red          & vintage      & natural      & new          & acrylic        \\
      bright        & purple       & purple       & romantic     & beautiful    & upper        & cold           \\
      blue          & colorful     & green        & deep         & delicate     & sole         & flip           \\
      welcome       & blue         & lace         & waist        & recycled     & genuine      & newborn        \\
      exact         & vibrant      & yellow       & front        & chic         & friendly     & large          \\
      yellow        & ruffle       & sweet        & gentle       & formal       & sexy         & floral         \\
      red           & orange       & french       & formal       & decorative   & stretchy     & waist          \\
      specific      & only         & black        & delicate     & romantic     & great        & american       \\
    \hline
    \end{tabular}
  \end{subtable}
  \begin{subtable}{\columnwidth}
    \centering
    \caption{Wear dataset}
    \begin{tabular}{|p{.132\columnwidth}|p{.132\columnwidth}|p{.132\columnwidth}|p{.132\columnwidth}|p{.132\columnwidth}|p{.132\columnwidth}|p{.132\columnwidth}|}
    \hline
    \bf norm1      & \bf norm2      & \bf conv3      & \bf conv4      & \bf pool5      & \bf fc6        & \bf fc7        \\
    \hline
    blue \newline green \newline red-black \newline red \newline denim-on-denim \newline denim-shirt \newline pink \newline denim \newline yellow \newline leopard &
    denim-jacket \newline pink \newline red \newline red-socks \newline red-black \newline champion \newline blue \newline white \newline shirt \newline i-am-clumsy \newline yellow &
    border-striped-tops \newline border-stripes \newline dark-style \newline stripes \newline backpack \newline red \newline dark-n-dark \newline denim-shirt \newline navy \newline outdoor-style &
    kids \newline bucket-hat \newline hat-n-glasses \newline black \newline sleeveless \newline american-casual \newline long-cardigan \newline white-n-white \newline stole \newline mom-style &
    shorts \newline half-length \newline pants \newline denim \newline dotted \newline border-stripes \newline white-pants \newline border-tops \newline gingham-check \newline sandals \newline chester-coat &
    white-skirt \newline flared-skirt \newline spring \newline upper \newline beret \newline shirt-dress \newline overalls \newline hair-band \newline loincloth-style \newline matched-pair &
    long-skirt \newline suit-style \newline midi-skirt \newline gaucho-pants \newline handmade \newline straw-hat \newline white-n-white \newline white-coordinate \newline white-pants \newline white \\
    \hline
    \end{tabular}
  \end{subtable}
  
\end{table}

\subsubsection{How fine-tuning affects perceptual depth}
\begin{figure}[t]
  \centering
  \begin{minipage}{.6\columnwidth}
    \includegraphics[width=\columnwidth]{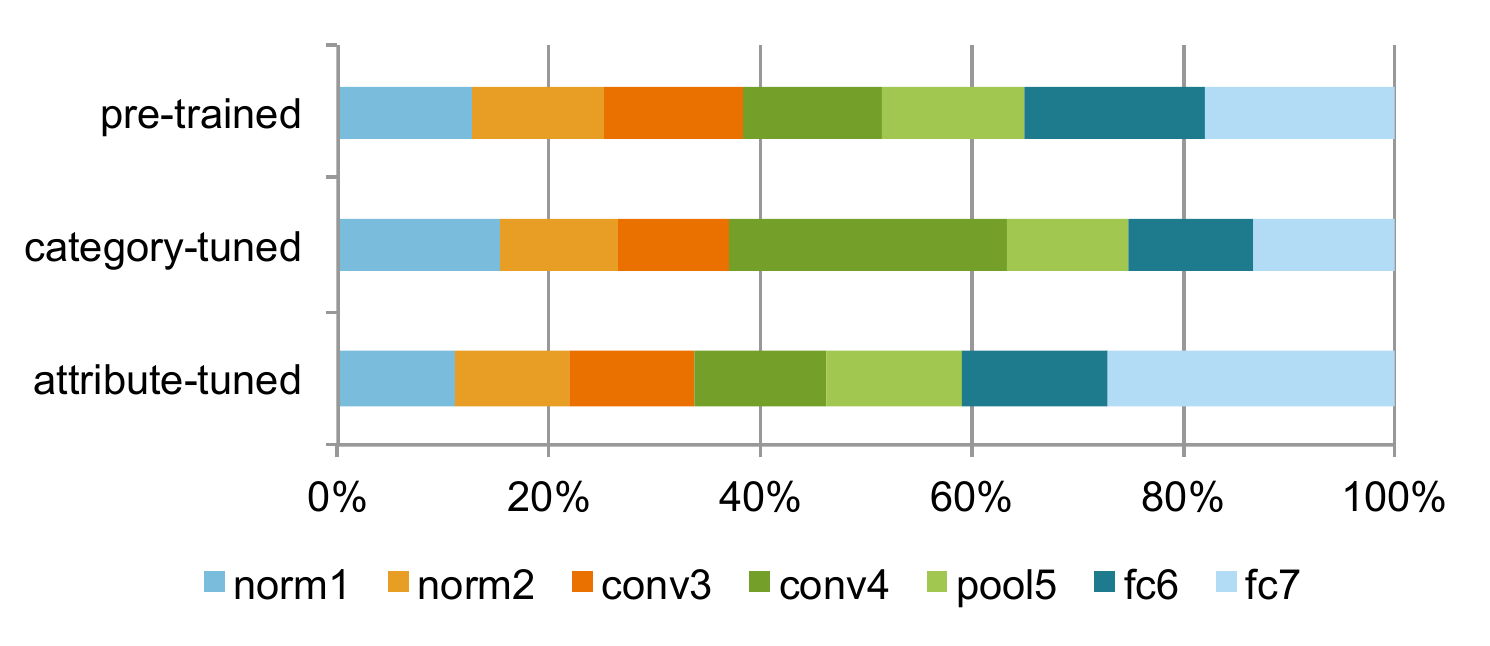}
    \caption{Relative magnitude of average layer-wise maximum KL divergence.}
    \label{fig:relative_average_maxkl}
  \end{minipage}
  \begin{minipage}{.38\columnwidth}
    \includegraphics[width=\columnwidth]{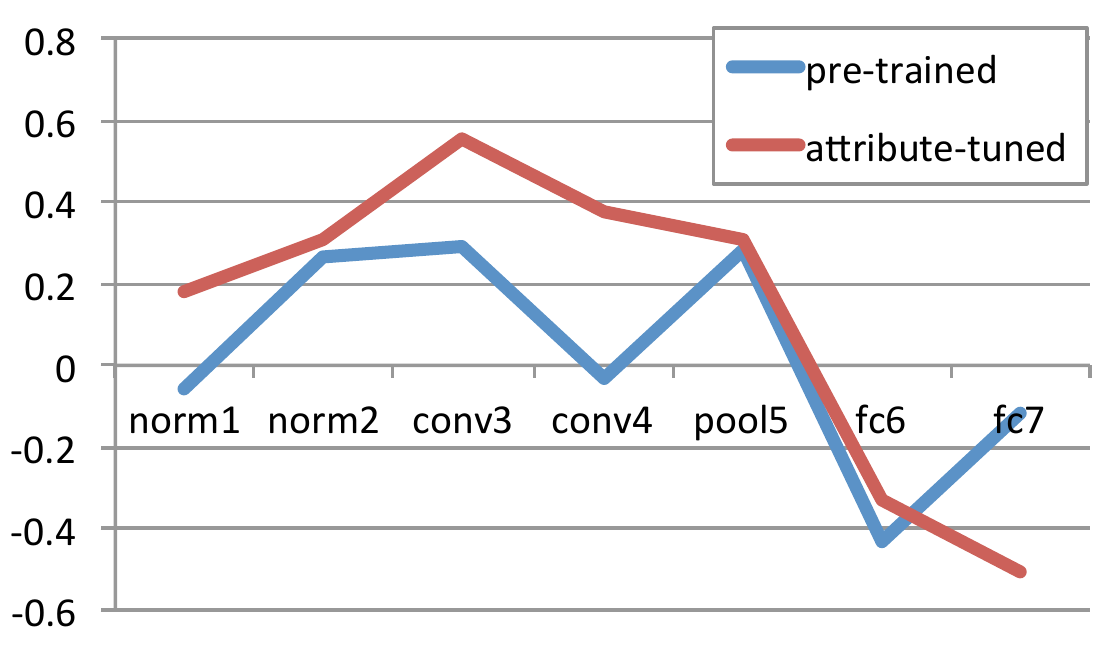}
    \caption{Pearson correlation coefficients between human visualness and max KL divergence of each CNN layer.}
    \label{fig:human_vs_layer}
  \end{minipage}
\end{figure}

Fine-tuning has an influence on the magnitude of the layer-wise max-pooled KL in that 1) the pre-trained model activates almost equally across layers and 2) the category-tuned model induced larger divergence in mid-layer (conv4), while 3) the attribute-tuned model activates more in the last layer (fc7). Figure \ref{fig:relative_average_maxkl} shows the relative magnitude of average layerwise max-pooled KL: $M_l \equiv \frac{1}{|U|} \sum_{u \in U} \sum_{i \in l} S_i(u|D)$. The attribute-tuning causes a direct change in the last layer as expected, whereas the category-tuning brings a representational change in the mid-layers. The result suggests the domain-specific knowledge is encoded inside the mid-to-higher level representation, but there are domain-agnostic features in the earliest layers perhaps useful for recognizing primitive attributes such as color.
Moreover, we also observe that the set of salient words per layer stay similar after fine-tuning in either cases; earlier layers activate more on primitive attributes, color or texture, and later layers activate more on abstract words.

\subsubsection{How each layer relates to human perception}
Finally, we evaluate how each layer relates to human perception, using the annotation from Sec \ref{sec:human_perception}. Figure \ref{fig:human_vs_layer} plots Pearson correlation of the layer-wise maximum KL divergence (eq \ref{eq:layerwise_kl}) against human visualness. We show the correlation of pre-trained and attribute-tuned CNNs. The plot suggests the activation of mid-layers is closer to human visualness perception, but interestingly, the last fully-connected layers give negative correlation. We think that the last layers are more associated to abstract words that are not generally considered visual by humans, but they are contextually associated in a domain-specific data as in {\it genuine leather} case.

\section{Saliency detection}


\subsubsection{Cumulative receptive fields}
We consider saliency detection with respect to the given attribute based on the receptive field~\cite{Zhou2014}. The main idea is to accumulate neurons' response in the order of the largest KL divergence. Following \cite{Zhou2014}, we first apply a sliding-window occluder to the given image, feed the occluded image to the CNN, and observe the difference in activation as a function of the occluder location $a_i(x, y)$ for unit $i$. We take the occluder patch at $(x, y)$ from the mean image of the dataset, at different scales. In our experiment, we use $24 \times 24$, $48 \times 48$, and $96 \times 96$ occluder size with stride size 4 for the $256 \times 256$ image input to the CNN. After getting the response map $a_i(x, y)$, we apply a Gaussian filter with the scale proportional to the occluder size, and average out multiple scale responses to produce a single response map. The resulting response map $A_i(x, y)$ can have either positive or negative peaks to the input pattern, and we heuristically negate and invert the response map if the map has negative peaks. We normalize the response map within $[0, 1]$ scale. Let us denote this normalized response map of unit $i$ by $R_i(x, y)$. We compute the final saliency map $M$ by accumulating units ordered and weighted by the KL divergence:
\begin{equation}
M(x, y|u, I) \equiv \frac{1}{Z} \sum_{i}^K S_i(u|D) R_i(x, y|I),
\end{equation}
where $Z = \sum_{i}^K S_i(u|D)$. We accumulate units by the largest unit divergence $S_i(u|D)$ up to $K$.

\subsubsection{Human annotation}
We use Wear dataset for saliency evaluation, since the images in Etsy dataset are mostly a single object appearing in the center of the image frame and there is merely a localization need. Similarly to Sec \ref{sec:human_perception}, we collect human annotation on the salient region for evaluation purpose. For the randomly selected set of 10 positive images for the most frequent 50 tags in Wear dataset, we ask 3 workers to draw bounding boxes around the relevant region to the specified tag-word. We consider pixels having 2 or more annotator votes to be the ground-truth salient regions. We discard images not having any worker agreement in the evaluation.

\subsubsection{Experimental results}
\begin{figure}[t]
  \centering
  \begin{subfigure}[b]{.43\columnwidth}
    \includegraphics[width=\columnwidth]{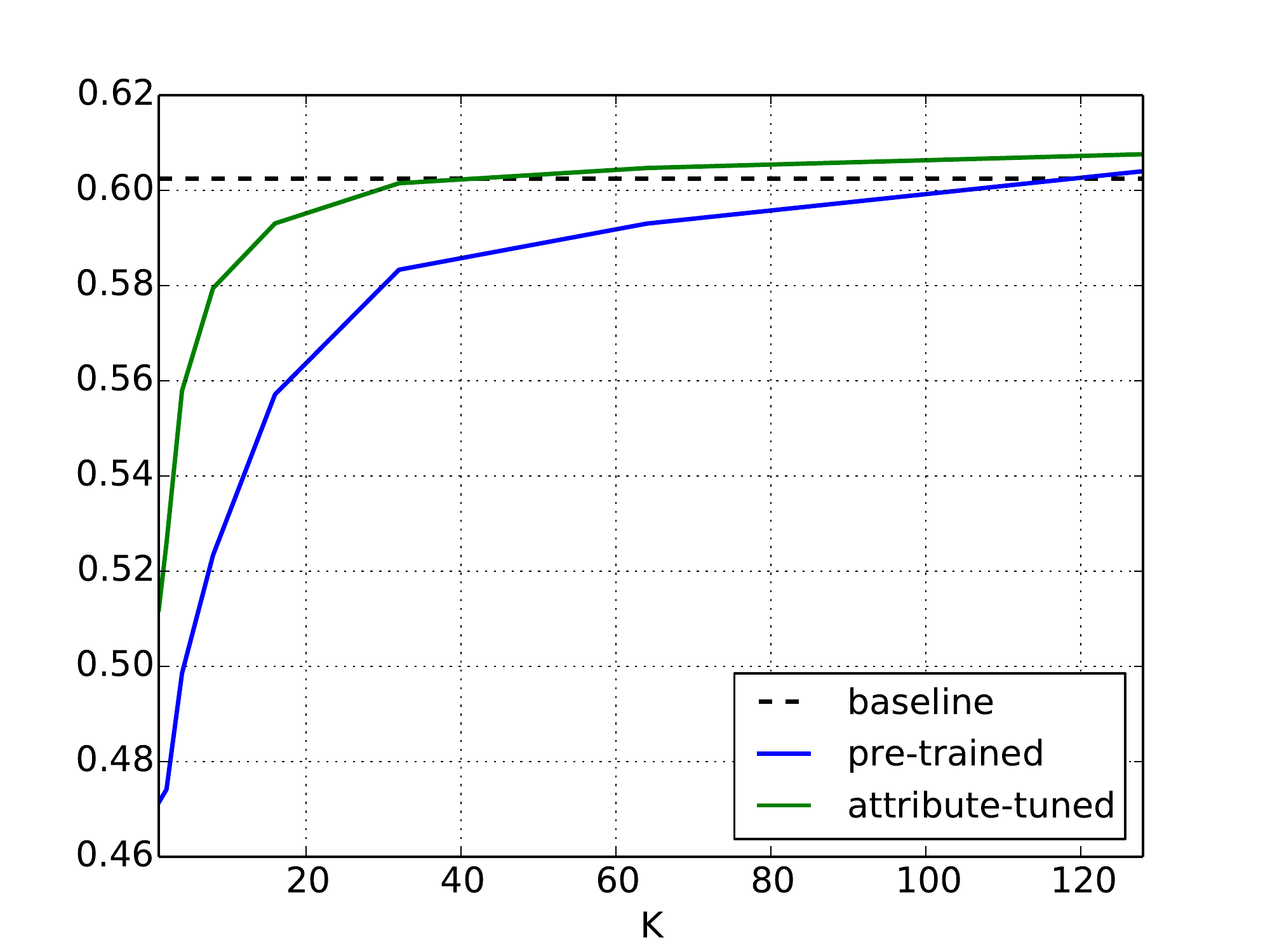}
    \caption{Mean AP}
    \label{fig:saliency_mAP}
  \end{subfigure}
  \begin{subfigure}[b]{.43\columnwidth}
    \includegraphics[width=\columnwidth]{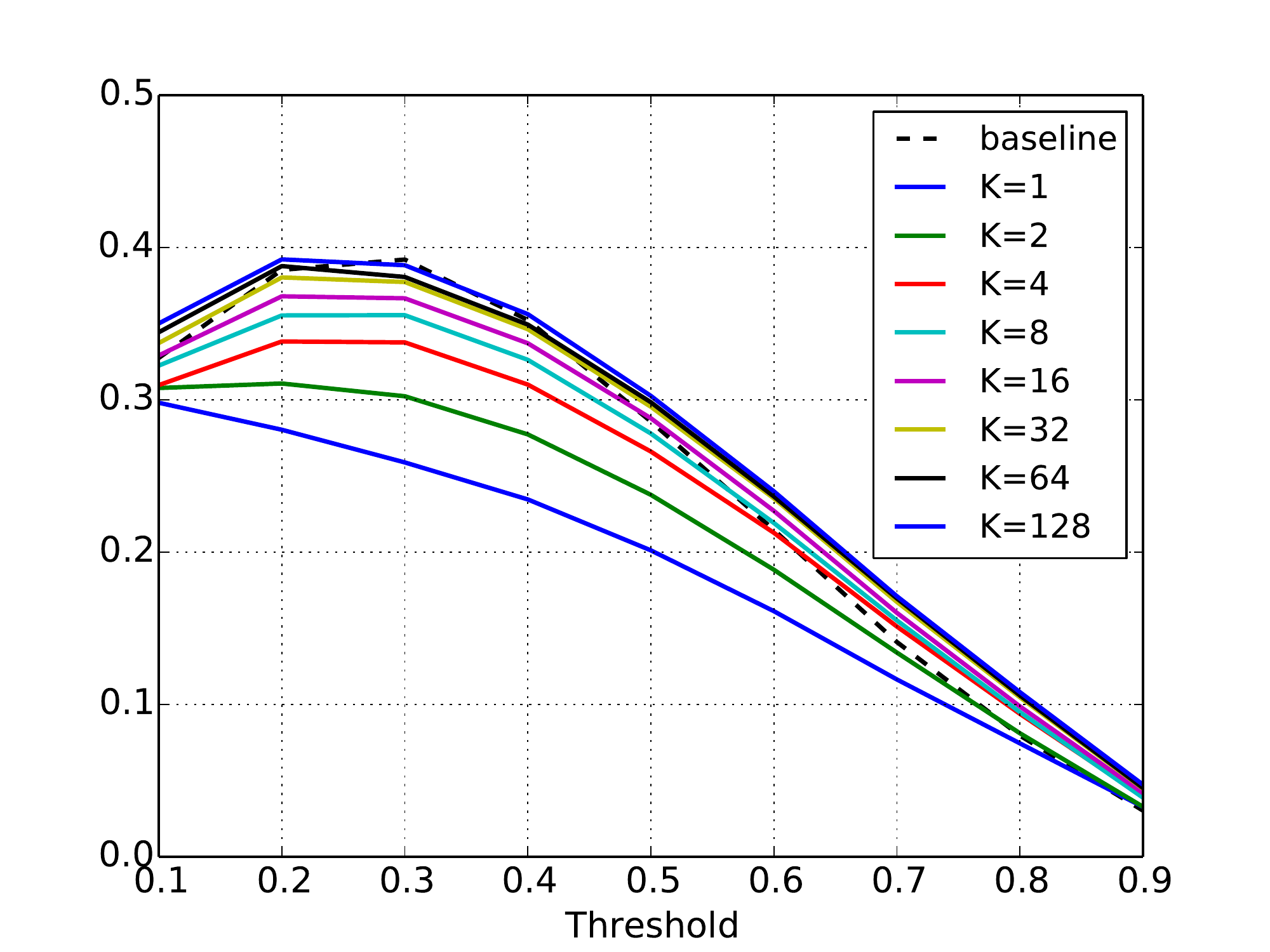}
    \caption{Mean IoU}
    \label{fig:saliency_iou}
  \end{subfigure}
  \caption{Saliency detection performance in terms of (a) mean average precision and (b) mean IoU of the attribute-tuned model over the heat-map threshold. 
  }
  \label{fig:saliency_evaluation}

\end{figure}

Figure \ref{fig:saliency_evaluation} plots the average performance from all the tags in terms of mean average precision (mAP) for predicting pixel-wise binary labels, and mean intersection-over-union (IoU) of the attribute-tuned model. We compute IoU for the binarized saliency map $M(x, y|u, I) \geq \theta$ at different threshold $\theta$. The plots show the performance with respect to the number of accumulations $K$, as well as the baseline performance of the smoothed gradient magnitude~\cite{Simonyan2014} of the attribute-tuned model. The performance improves as more neurons accumulate in the saliency map according to the divergence, and gives on par or slightly better performance against the baseline. Note that even the pre-trained model can already reach the baseline by this simple accumulation based on KL divergence, without any optimization towards saliency. We observe improvement in both pre-trained and attribute-tuned models, but the pre-trained model tends to require more neurons. We believe that fine-tuning makes each neuron activate more to a specific pattern while reducing activations on irrelevant patterns, and that results in the diminishing accumulation effect. The result also suggests that visual attributes are combinatorial visual stimuli rather than some visual pattern detectable only with a single neuron.

\begin{figure}[t]
  \centering
  \includegraphics[width=\columnwidth]{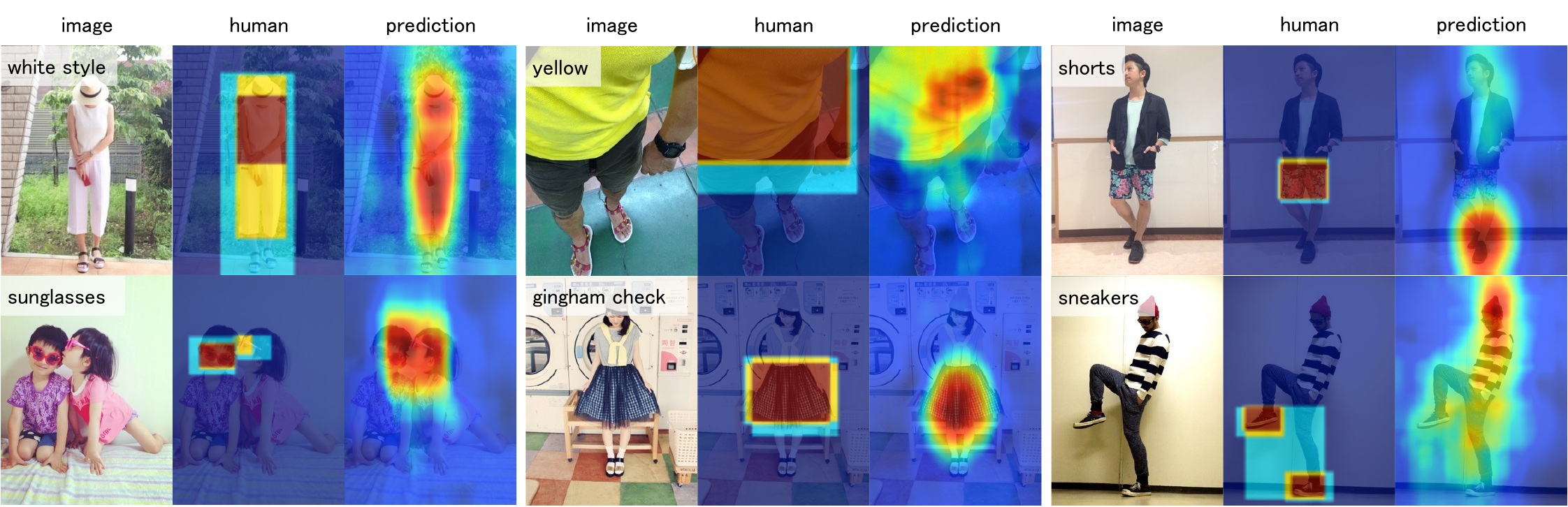}
  \caption{Results of detected salient regions for the given attribute. The rightmost column shows failure cases due to distracting contexts or visibility issues.}
  \label{fig:saliency_examples}
\end{figure}
\begin{figure}[t]
  \centering
  
  \includegraphics[width=.85\columnwidth]{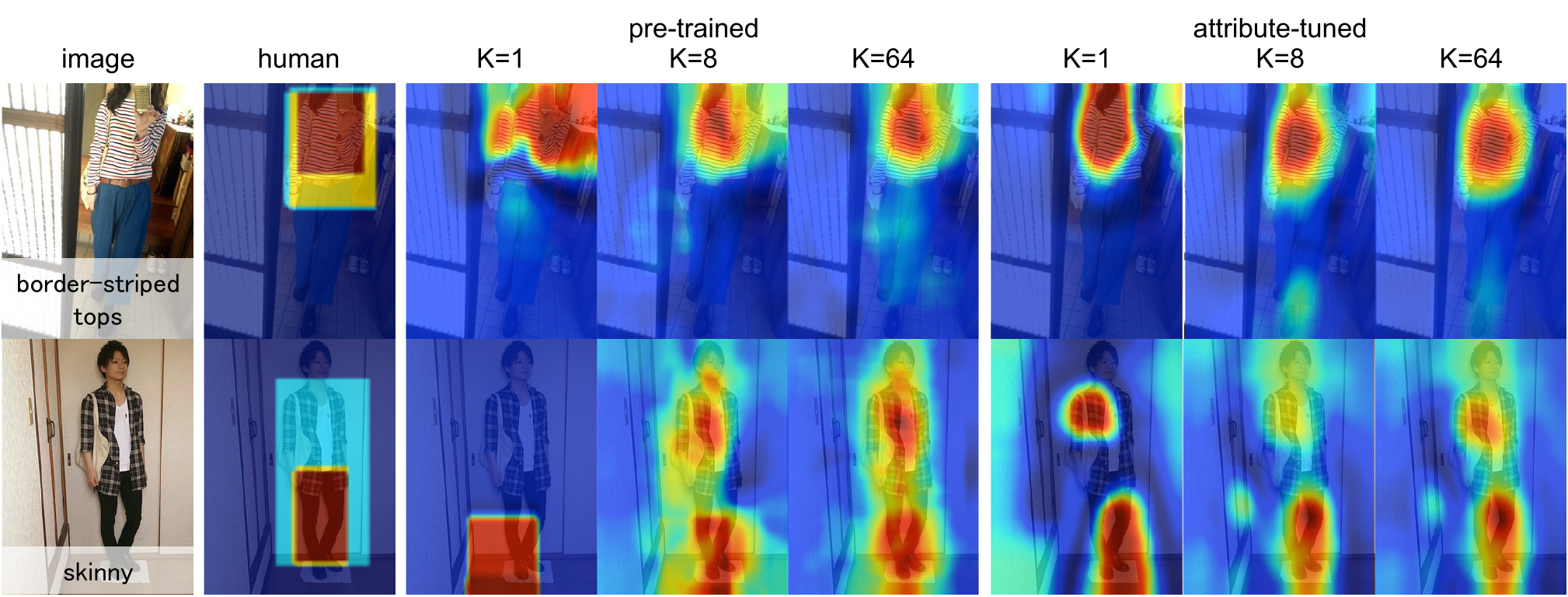}
  \caption{Accumulating receptive fields by the largest KL divergence. As we add more neurons, the saliency heat-map becomes finer.}
  \label{fig:cumulative_rf}
  
\end{figure}

Figure \ref{fig:cumulative_rf} shows the detection results by human annotation and our cumulative receptive field using the pre-trained and fine-tuned CNNs, when the accumulation size $K$ is 1, 8, and 64. Also, Figure \ref{fig:saliency_examples} shows the results of human annotation and the pre-trained CNN with accumulation size $K = 64$. Our saliency detection method works remarkably well \emph{even without fine-tuning}. As we accumulate more neurons, the response map tends to produce finer localization. Accumulation helps most of the cases, but we observe failure cases when there is a distractor co-occuring with the given attribute. For example in Figure \ref{fig:saliency_examples}, detecting shorts fails because legs always appear with shorts and we end up with legs detector instead of shorts (distractor issue). Moreover, our method tends to fail when the target attribute is associated to only small regions in the image (visibility issue). 

\section{Conclusion}
We have shown that we are able to discover and analyze a new visual attribute from noisy Web data using neural activations. The key idea is the use of highly activating neurons in the network, identified by the divergence of activation distribution in the weakly annotated dataset. Empirical study using two real-world data gives us insights that our approach can automatically learn a visual attribute classifier that has a perceptual ability similar to humans, the depth in the network relates to the depth of attribute perception, and the neurons can detect salient regions in the given image. In the future, we wish to further study the relationship and similarity between discovered visual attributes and how the network architecture changes the neural perception in the hierarchical structure.

\subsubsection{Acknowledgement}
This work was partly supported by JSPS KAKENHI Grant Number 15H05919 

\bibliographystyle{splncs}
\bibliography{main}
\end{document}